\documentclass{article}

\usepackage{graphicx}
\usepackage{xspace}
\usepackage{amsmath}
\newcommand{\BfPara}[1]{\noindent\textbf{#1.}\xspace}

\usepackage[dblblindworkshop, final]{neurips_2026}
 \workshoptitle{SaTQuML: Secure and Trustworthy Quantum Machine Learning}

\usepackage[utf8]{inputenc} 
\usepackage[T1]{fontenc}    
\usepackage{hyperref}       
\usepackage{url}            
\usepackage{booktabs}       
\usepackage{amsfonts}       
\usepackage{nicefrac}       
\usepackage{microtype}      
\usepackage{xcolor}         

\title{vFedProtoQNAS: Prototype-Guided Personalized Quantum Neural Architecture Search for Virtual Federated Learning}

\author{%
  Seok Bin Son \\
  Korea University \\
  Seoul, Korea \\
  \texttt{lydiasb@korea.ac.kr}
  \And
  Samuel Yen-Chi Chen\\
  Wells Fargo \\
  New York, USA \\
  \texttt{ycchen1989@ieee.org}
  \AND
  Soohyun Park\thanks{Corresponding authors: Soohyun Park and Joongheon Kim.} \\
  Sookmyung Women's University \\
  Seoul, Korea \\
  \texttt{soohyun.park@sookmyung.ac.kr}
  \And
  Joongheon Kim$^{*}$ \\
  Korea University \\
  Seoul, Korea \\
  \texttt{joongheon@korea.ac.kr}
}

\begin{document}

\maketitle

\begin{abstract}
Quantum federated learning (QFL) has emerged as a promising approach for collaboratively training compact quantum neural networks (QNNs) over distributed private data on resource-constrained devices. However, differences in device capabilities make a single shared QNN architecture unsuitable for all clients. While personalized quantum neural architecture search (QNAS) allows each client to select a device-specific QNN, averaging parameters across structurally different QNN architectures mixes semantically inconsistent circuit operations. To address this, prototype-guided personalized QNAS for virtual FL (vFedProtoQNAS) is proposed, where model parameters are never aggregated across clients and federated collaboration is achieved through class-wise prototype sharing. Each client independently searches and trains a client-specific QNN, computes class-wise local prototypes from latent representations, and refines them using global prototypes from the server as federated semantic anchors. Experiments demonstrate that vFedProtoQNAS improves accuracy by 3.70\% over FedAvg and enhances class-consistent representation alignment.
\end{abstract}

\section{Introduction}

Federated learning (FL) is a distributed learning paradigm that enables multiple clients to collaboratively train a model while preserving local data privacy~\citep{FL_1, DBLP:conf/cikm/KangP25}. This makes FL suitable for privacy-sensitive and communication-efficient applications, such as mobile devices, edge devices, and internet of things (IoT) networks. Most FL methods assume that clients share the same model architecture, allowing the server to construct a global model by averaging local parameters~\citep{QFL_1}. However, this assumption is often unrealistic in practical FL environments, where clients differ in computational resources, memory capacity, and energy budget \cite{10918773,CVPR_1,DBLP:conf/eccv/KhareAABLLT24,horvath2021fjord,mei2022resourceadaptive}. Such device heterogeneity makes it difficult to deploy a single unified architecture across all clients \cite{DBLP:journals/ict-express/ParkK24,alam2022fedrolex,yi2024fedmrl}. A high-capacity architecture may exceed the resource budget of low-resource clients, whereas an overly lightweight architecture may limit the learning capability of resource-rich clients. Therefore, heterogeneous FL requires compact, expressive, and adaptable model architectures that reflect client-specific capabilities, motivating resource-aware architecture design for distributed learning across diverse devices.

\begin{figure}[t!]
\centering
\includegraphics[width=0.7\linewidth]{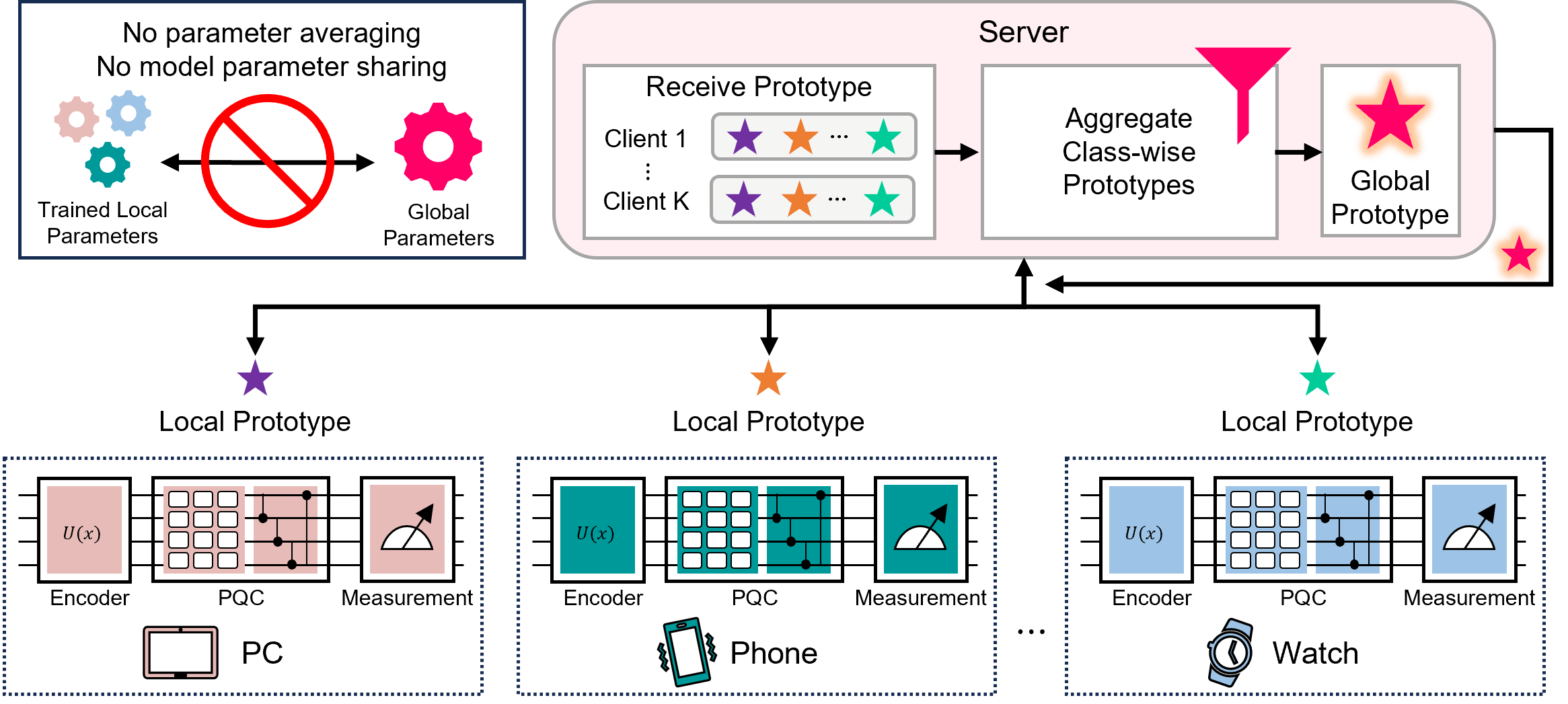}
\caption{The concept of virtual federated learning.}\label{fig:vfl}
\vspace{-5mm}
\end{figure}

Quantum neural networks (QNNs) are a potential compact model structure for distributed learning~\citep{QFL_1}. QNNs efficiently transform classical inputs into quantum feature spaces. Integrating QNNs with FL yields quantum FL (QFL), a promising approach for resource-constrained distributed setups~\citep{10296523}. However, deploying identical QNN architectures across clients is often unsuitable. Device-specific factors, such as circuit depth and gate type, affect performance and efficiency~\citep{QNAS_CIKM, AAAI_QNAS}. High-performance devices benefit from complex QNNs, while limited devices require simpler ones, necessitating client-specific architecture design. Manual design is challenged by interacting factors. Neural architecture search (NAS), and specifically quantum neural architecture search (QNAS), automates this process, enabling client-specific QNNs~\citep{QNAS_CIKM}. Applying personalized QNAS to QFL addresses the architecture mismatch and allows each client to select an optimal QNN based on device constraints and local data~\citep{DBLP:conf/aaai/YaoL24}. However, personalized QNAS complicates aggregation: FedAvg assumes uniform models and aligns client parameters, enabling meaningful averaging~\citep{DBLP:conf/ijcai/Yi0S00CL24}. Personalized QNAS breaks this alignment, so parameter indices across clients may represent different operations. Averaging without structural alignment mixes inconsistent information, undermining the specialization gained from personalized QNAS.

\begin{figure}[t!]
\centering
\includegraphics[width=0.7\linewidth]{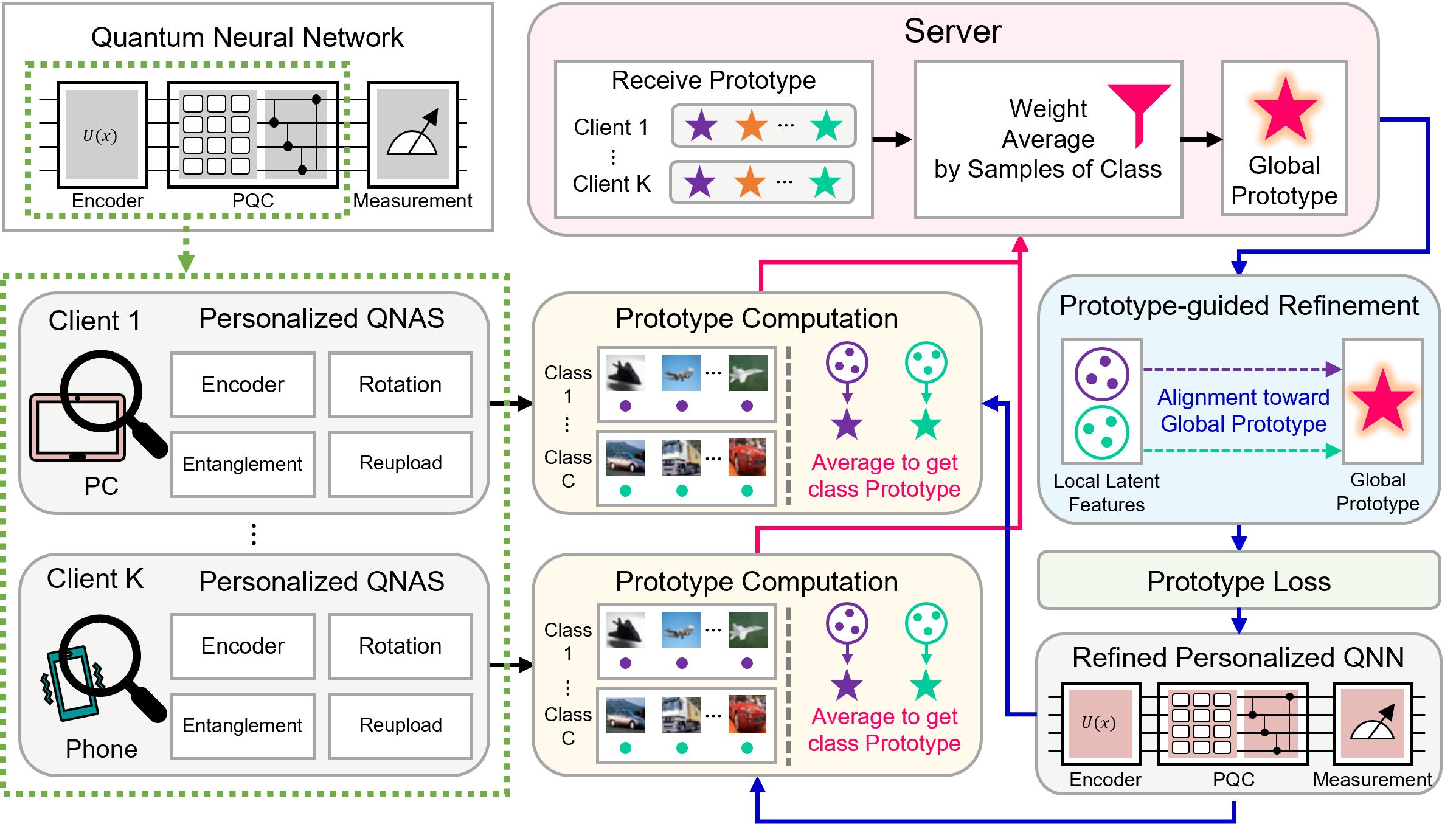}
\caption{The overall framework of vFedProtoQNAS.}\label{fig:overall_final}
\vspace{-5mm}
\end{figure}

To address this issue, this paper proposes vFedProtoQNAS: a prototype-guided, personalized QNAS framework for virtual FL. Unlike conventional FL, virtual here means model parameters are not aggregated, as shown in Fig.~\ref{fig:vfl}. vFedProtoQNAS shares class-wise prototypes for representation-level federated collaboration. Clients retain client-specific QNN architectures and model parameters locally, and collaborate with the server via prototype-based sharing. As shown in Fig.~\ref{fig:overall_final}, each client performs personalized QNAS tailored to its device constraints and local data, selects a QNN architecture, and trains its parameters. Each trained QNN maps samples to latent features, from which class-wise local prototypes are computed. These prototypes summarize class-level representations and are shared with the server \cite{tan2022fedpcl,zhang2024fedgmkd}. The server aggregates them using sample weights to generate a global prototype for each class, serving as a federated semantic anchor that integrates class-level representations across clients. The server sends global prototypes back. Clients refine their representations, encouraging alignment with the global prototype for the target class and separation from other classes. Client architectures and parameters are not averaged, preserving local specialization from personalized QNAS. Therefore, vFedProtoQNAS enables client-specific architecture search and federated collaboration in heterogeneous QFL, sharing knowledge at the representation level across different architectures.

The main contributions of this paper are as follows. (i) This paper proposes vFedProtoQNAS, a personalized QNAS framework that accounts for client-specific device constraints and local data characteristics in heterogeneous QFL. (ii) To mitigate structural mismatches between different QNN architectures, vFedProtoQNAS introduces a virtual federated collaboration method based on class-wise prototype sharing instead of parameter averaging. (iii) Through global prototype-based refinement, vFedProtoQNAS induces class-consistent representation alignment and achieves improved classification performance over major baselines.

\section{Preliminary}

\BfPara{QNN}
A QNN consists of an encoder, a parameterized quantum circuit (PQC), and a measurement stage. The encoder maps a classical input $x$ into a quantum representation using data-dependent rotation gates $U(x)$~\citep{QCNN_1}. The PQC applies trainable operations parameterized by $\theta$, where 1-qubit gates perform local rotations and 2-qubit gates generate entanglement. The measurement stage produces observables, such as Pauli-$Z$ expectations $\langle O\rangle_{x,\theta}\in[-1,1]^{|y|}$, which are used for classical readout. The gradient $\nabla_{\theta}\mathcal{L}(\theta)$ is evaluated using the parameter-shift rule \cite{mitarai2018quantum}, and the circuit parameters are updated as $\theta \leftarrow \theta - \eta \nabla_{\theta}\mathcal{L}(\theta)$.

\BfPara{QNAS}
QNN performance strongly depends on PQC design, including gate type, gate placement, circuit depth, and entanglement topology. More expressive circuits can improve representation capacity but also increase cost, noise sensitivity, and trainability issues. QNAS automates this design by searching over candidate circuit architectures \cite{ICML_QuantumDARTS,EPJQT_225,lu2023qasbench}. To handle the rapidly growing search space over qubits, layers, and operations, one-shot NAS trains a shared supernet and evaluates candidates through weight sharing.

\BfPara{QFL} QFL integrates FL with QNNs, enabling clients to collaboratively train QNNs without sharing local data \cite{9641742,10296523, chen2021federated,chehimi2022quantum}. Each client trains a local QNN and uploads its parameters to the server, which constructs a global QNN by averaging them as $\tilde{\theta}^{G} \leftarrow \frac{1}{K'} \sum_{k=1}^{K} c_k \tilde{\theta}^{k}$, where $\tilde{\theta}^{k}$ denotes the local QNN parameters, $c_k$ indicates client participation, and $K'=\sum_{k=1}^{K}c_k$. This parameter averaging assumes that all clients share the same QNN architecture and parameter space; otherwise, the same parameter index may correspond to different gates or circuit paths, making aggregation inconsistent.

\vspace{-3mm}

\section{vFedProtoQNAS Framework}

\subsection{Problem Formulation}
This paper considers a heterogeneous QFL environment with $K$ clients. 
Each client $i \in \{1,\ldots,K\}$ owns a private local dataset 
$\mathcal{D}_i$, which is not shared with other clients or the server. 
The overall data distribution is non-IID across clients. Clients may have 
different computational resources, memory capacities, supported circuit 
depths, and executable quantum operations. Therefore, the feasible QNN 
architecture space can vary across clients. The architecture search space 
of client $i$ is denoted by $\mathcal{A}_i$. Although the internal QNN architectures may differ, vFedProtoQNAS constrains all clients
to use a common latent interface defined by the same number and ordering of measured observables
for prototype computation. QNAS changes the internal PQC structure while preserving this common
output interface, allowing class-wise prototypes to be represented and aggregated in the same
measurement-defined latent space. The objective of vFedProtoQNAS is to identify a personalized QNN architecture 
that reflects both the local data distribution and the device constraints of 
each client. Accordingly, the client-specific joint optimization problem is 
formulated as follows:
\begin{equation}
\min_{\substack{\alpha_i \in \mathcal{A}_i\\ \theta_i}}
\left[
\mathcal{L}_{\mathrm{CE}}^{i}
(\theta_i,\alpha_i;\mathcal{D}_i)
+
\lambda C(\alpha_i)
\right].
\label{eq:local_objective}
\end{equation}

In Eq.~\eqref{eq:local_objective}, $\alpha_i$ and $\theta_i$ denote the 
architecture and model parameters of client $i$, respectively. 
$\mathcal{L}_{\mathrm{CE}}^{i}$ denotes the local classification loss, 
whereas $C(\alpha_i)$ represents the resource cost of the selected 
architecture. The coefficient $\lambda$ controls the trade-off between 
classification performance and architecture cost. Personalized QNAS, however, introduces a fundamental aggregation challenge. 
FedAvg assumes that all clients share the same model architecture and that 
their parameters belong to an aligned parameter space. This assumption does 
not hold when clients select different QNN architectures. The same parameter 
index may correspond to different quantum operations, circuit depths, or 
representation paths. Direct parameter averaging can therefore mix 
structurally misaligned information. To address this problem, vFedProtoQNAS 
replaces parameter-level aggregation with representation-level collaboration 
based on class-wise prototypes.

\subsection{Architecture}
As shown in Fig.~\ref{fig:overall_final}, vFedProtoQNAS proceeds sequentially through personalized architecture search,
local training, prototype aggregation, and prototype-guided refinement. Each client first performs
personalized QNAS within $\mathcal{A}_i$, fixes the selected architecture $\alpha_i$, and trains its
parameters $\theta_i$ using local data. The trained QNN then computes class-wise local prototypes,
which are sent with the corresponding class sample counts to the server while raw data and model
parameters remain local. The server aggregates them into global prototypes and returns them to the
clients. Each client finally refines $\theta_i$ using the global prototypes while keeping $\alpha_i$
fixed, thereby preserving personalized architectures and local specialization.

\subsection{Client-Specific ProxylessQNAS}
Each client performs architecture search within its feasible space $\mathcal{A}_i$, whose maximum
circuit depth and available operations are determined by its device constraints. In our experiments,
the search space includes rotation, entanglement, and data re-uploading operations over up to four
layers. vFedProtoQNAS adopts a ProxylessQNAS strategy in which each candidate operation
$g \in \mathcal{G}_m$ at searchable layer $m$ is associated with an architecture logit $z_{m,g}$.
\begin{equation}
\pi_{m,g}
=
\frac{\exp(z_{m,g})}
{\sum_{g' \in \mathcal{G}_m}\exp(z_{m,g'})}.
\label{eq:operation_probability}
\end{equation}

In Eq.~\eqref{eq:operation_probability}, $\pi_{m,g}$ represents the 
probability of selecting operation $g$ at layer $m$. The architecture logits are optimized using the local validation objective so that candidate
operations with better validation performance receive higher selection probabilities. These
probabilities guide the cost-aware architecture search according to
\begin{equation}
\min_{\alpha_i \in \mathcal{A}_i}
\left[
\mathcal{L}_{\mathrm{val}}^{i}(\alpha_i,\theta_i)
+
\lambda_{\mathrm{cost}} C(\alpha_i)
\right].
\label{eq:search_objective}
\end{equation}
In Eq.~\eqref{eq:search_objective}, 
$\mathcal{L}_{\mathrm{val}}^{i}$ denotes the validation loss of client $i$, 
and $C(\alpha_i)$ represents the resource cost of architecture $\alpha_i$. 
The coefficient $\lambda_{\mathrm{cost}}$ balances predictive performance 
and resource efficiency. Consequently, each client selects an executable QNN 
architecture optimized within its own feasible search space.

\subsection{Prototype Computation and Aggregation}
After selecting the client-specific architecture $\alpha_i$, client $i$ fixes 
the architecture and trains its model parameters $\theta_i$ using 
$\mathcal{D}_i$. The trained QNN maps an input sample $x$ to a latent 
representation $h_i(x)$. Let $\mathcal{D}_{i,c}$ denote the subset of local 
samples belonging to class $c$, and let 
$n_{i,c}=|\mathcal{D}_{i,c}|$.

For every locally observed class satisfying $n_{i,c}>0$, the class-wise local 
prototype is computed as:
\begin{equation}
p_{i,c}
=
\frac{1}{n_{i,c}}
\sum_{(x,y)\in\mathcal{D}_{i,c}} h_i(x).
\label{eq:local_prototype}
\end{equation}

In Eq.~\eqref{eq:local_prototype}, $p_{i,c}$ represents the local center of class $c$ in the common latent space. Each client 
transmits only the prototypes of its observed classes and the corresponding 
sample counts to the server. Raw data and model parameters are not 
transmitted. To handle classes that may be absent from some clients under non-IID data, 
let $\mathcal{K}_c=\{i \mid n_{i,c}>0\}$ denote the set of clients containing 
at least one sample of class $c$. For each class represented by at least one 
client, the server computes the global prototype as:
\begin{equation}
\bar{p}_c
=
\frac{
\sum_{i\in\mathcal{K}_c} n_{i,c}p_{i,c}
}{
\sum_{i\in\mathcal{K}_c} n_{i,c}
}.
\label{eq:global_prototype}
\end{equation}

Equation~\eqref{eq:global_prototype} performs sample-weighted aggregation, 
assigning greater influence to clients with more samples of class $c$. The 
resulting global prototype $\bar{p}_c$ serves as a federated semantic anchor 
that integrates class-level knowledge across heterogeneous client models.

\subsection{Prototype-Guided Local Refinement}
After prototype aggregation, the server returns the global prototype set to 
the clients. Each client uses these prototypes to refine its personalized QNN 
by jointly preserving local classification performance and aligning local 
representations with global class-level semantics. The refinement objective 
of client $i$ is defined as:
\begin{equation}
\mathcal{L}_i
=
\mathcal{L}_{\mathrm{CE}}^{i}
+
\beta_{\mathrm{proto}}
\mathcal{L}_{\mathrm{proto}}^{i}.
\label{eq:refinement_objective}
\end{equation}

In Eq.~\eqref{eq:refinement_objective}, 
$\mathcal{L}_{\mathrm{CE}}^{i}$ preserves the local discriminative ability of 
the personalized QNN, while $\mathcal{L}_{\mathrm{proto}}^{i}$ promotes 
alignment with the global prototypes. The coefficient 
$\beta_{\mathrm{proto}}$ controls the strength of prototype guidance. The prototype alignment loss is obtained by averaging a contrastive 
per-sample loss over the local training data. For a sample $x$ with 
ground-truth label $y$, the per-sample loss is defined as:
\begin{equation}
\ell_{\mathrm{proto}}^{i}(x,y)
=
-\log
\frac{
\exp\!\left(
\mathrm{sim}(h_i(x),\bar{p}_y)/\tau
\right)
}{
\sum_{c=1}^{C}
\exp\!\left(
\mathrm{sim}(h_i(x),\bar{p}_c)/\tau
\right)
}.
\label{eq:prototype_loss}
\end{equation}

In Eq.~\eqref{eq:prototype_loss}, $\mathrm{sim}(\cdot,\cdot)$ denotes cosine 
similarity, and $\tau$ is the temperature parameter. The numerator measures 
the similarity between the local representation and the global prototype of 
the correct class, while the denominator compares it with all class 
prototypes. This objective pulls the local representation toward the correct 
global prototype and separates it from incorrect prototypes. During 
refinement, each client updates only its local parameters $\theta_i$, while 
the personalized architecture $\alpha_i$ remains fixed. Therefore, 
vFedProtoQNAS enables federated knowledge sharing through class-consistent 
representation alignment without aggregating structurally incompatible 
model parameters.

\section{Performance Evaluation}
\subsection{Simulation Setup} The experiment was conducted on the MNIST $(0$–$3)$ four-class classification task. Non-IID data were distributed across five clients using a Dirichlet distribution. Each QNN used four latent tokens and up to four layers. The operation set and circuit depth were configured based on client-specific device constraints. All experiments were repeated over five random seeds, and the results are reported as mean $\pm$ standard deviation. Each client was locally trained for 100 epochs, followed by one prototype aggregation step and a
100-epoch prototype-guided refinement phase. Prototype aggregation and refinement were performed
once rather than over repeated federated communication rounds. Contrastive prototype loss was applied with $\beta_{\mathrm{proto}}=0.10$. Adam was used as the optimizer with a batch size of 64.

\begin{figure}[t!]
    \centering
    \includegraphics[width=0.7\columnwidth]
    {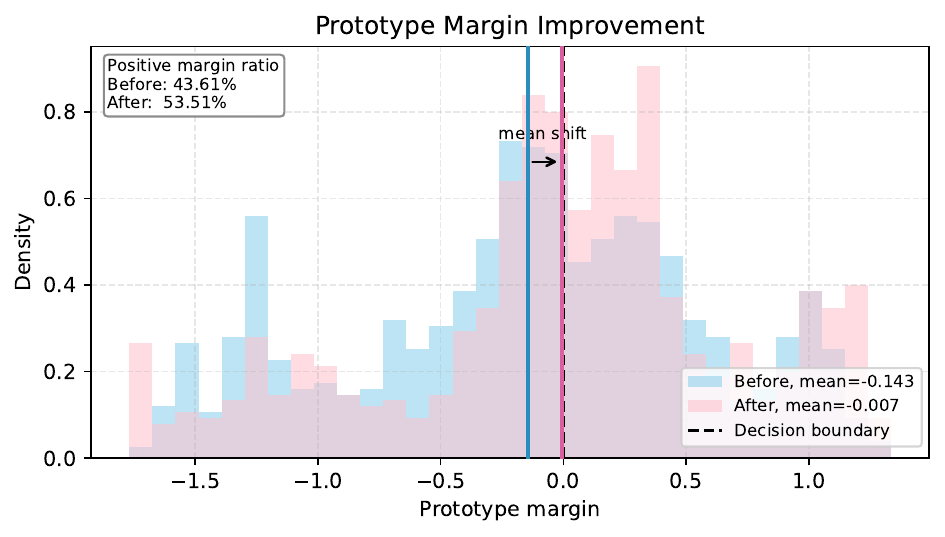}
    \caption{Prototype margin improvement before and after prototype-guided refinement. A positive margin indicates that a latent representation is closer to the correct class prototype than to any incorrect prototype.}
    \label{fig:proto_margin}
    \vspace{-3mm}
\end{figure}

\subsection{Experimental Results}

\BfPara{Prototype Margin Analysis}
To evaluate whether prototype-guided refinement induces class-consistent representation alignment, the prototype margin is analyzed. The prototype margin of a sample $x$ is defined as $m(x) = \mathrm{sim}(h_i(x),\bar{p}_{y}) - \max_{c\neq y} \mathrm{sim}(h_i(x),\bar{p}_{c}).$ This margin measures the similarity gap between the correct class prototype and the closest incorrect prototype. A positive margin indicates that $h_i(x)$ is closer to the correct prototype than to any incorrect prototype. Therefore, the prototype margin provides a direct measure of class-consistent representation alignment. Fig.~\ref{fig:proto_margin} compares the margin distributions before and after prototype-guided refinement. After refinement, the distribution shifts toward the positive region. The mean margin improves from $-0.143$ to $-0.007$. The positive margin ratio also increases from $43.61\%$ to $53.51\%$. These results indicate that vFedProtoQNAS improves class-consistent alignment of client-specific representations.

\begin{table}[t]
\centering
\small
\caption{Sensitivity analysis of vFedProtoQNAS with respect to the prototype alignment weight $\beta_{\mathrm{proto}}$ using MNIST(0-3). Results are reported as mean$\pm$std over five seeds.}
\label{tab:beta_sensitivity}
\setlength{\tabcolsep}{5pt}
\renewcommand{\arraystretch}{1.05}
\begin{tabular}{c|cccc}
\toprule
$\beta_{\mathrm{proto}}$ 
& Accuracy 
& Precision 
& Recall 
& F1-score \\
\midrule
$0.05$ 
& $\mathbf{91.25{\pm}0.53}$ 
& $\mathbf{91.30{\pm}0.52}$ 
& $\mathbf{91.25{\pm}0.53}$ 
& $91.23{\pm}0.54$ \\

$0.10$ 
& $\mathbf{91.25{\pm}0.74}$ 
& $91.29{\pm}0.71$ 
& $\mathbf{91.25{\pm}0.74}$ 
& $\mathbf{91.24{\pm}0.74}$ \\

$0.20$ 
& $91.19{\pm}0.66$ 
& $91.23{\pm}0.66$ 
& $91.19{\pm}0.66$ 
& $91.18{\pm}0.68$ \\

$0.50$ 
& $90.94{\pm}0.38$ 
& $90.98{\pm}0.35$ 
& $90.94{\pm}0.38$ 
& $90.93{\pm}0.37$ \\
\bottomrule
\end{tabular}
\vspace{-5mm}
\end{table}

\BfPara{Sensitivity Analysis of the Prototype Alignment Weight} Table~\ref{tab:beta_sensitivity} reports the sensitivity of vFedProtoQNAS to the prototype alignment weight $\beta_{\mathrm{proto}}$. vFedProtoQNAS shows stable performance under small and moderate values of $\beta_{\mathrm{proto}}$. For $\beta_{\mathrm{proto}}=0.05$ and $\beta_{\mathrm{proto}}=0.10$, the accuracy reaches $91.25\pm0.53\%$ and $91.25\pm0.74\%$, respectively. The F1-score also remains stable at $91.23\pm0.54\%$ and $91.24\pm0.74\%$, respectively. This result indicates that moderate prototype guidance improves class-wise representation alignment without disrupting local classification learning. In contrast, when $\beta_{\mathrm{proto}}=0.50$, the accuracy decreases to $90.94\pm0.38\%$. The F1-score also decreases to $90.93\pm0.37\%$. This result shows that the benefit of prototype guidance does not increase monotonically with its weight. The prototype alignment term acts as a representation regularizer based on global prototypes. It does not replace the local classification objective. Therefore, an excessively large $\beta_{\mathrm{proto}}$ can weaken local specialization. This occurs because client-specific representations are overly constrained toward the global prototype space. Overall, these results show that prototype guidance should be applied with a moderate weight. Its role is to balance local discriminative learning and global representation alignment. It is not intended to enforce excessive representation alignment.

\begin{table}[t]
\centering
\small
\caption{Sensitivity analysis of vFedProtoQNAS under different non-IID levels using MNIST(0-3). Results are reported as mean$\pm$std over five seeds. A smaller Dirichlet concentration parameter $\alpha$ indicates stronger data heterogeneity.}
\label{tab:alpha_sensitivity}
\setlength{\tabcolsep}{5pt}
\renewcommand{\arraystretch}{1.05}
\begin{tabular}{c|cccc}
\toprule
$\alpha$
& Accuracy
& Precision
& Recall
& F1-score \\
\midrule
$0.5$
& $84.32{\pm}11.77$
& $82.67{\pm}16.72$
& $84.32{\pm}11.77$
& $82.64{\pm}15.48$ \\

$1.0$
& $91.15{\pm}0.76$
& $\mathbf{91.26{\pm}0.67}$
& $91.15{\pm}0.76$
& $91.15{\pm}0.75$ \\

$10.0$
& $\mathbf{91.16{\pm}0.93}$
& $91.24{\pm}0.89$
& $\mathbf{91.16{\pm}0.93}$
& $\mathbf{91.17{\pm}0.92}$ \\
\bottomrule
\end{tabular}
\vspace{-3mm}
\end{table}

\begin{table}[t!]
\centering
\small
\caption{Comparison with baselines under the default setting $\beta_{\mathrm{proto}}=0.10$. Results are reported as mean$\pm$std over five seeds.}
\label{tab:main_comparison}
\setlength{\tabcolsep}{4pt}
\renewcommand{\arraystretch}{1.05}
\begin{tabular}{l|cccc}
\toprule
\textbf{Method}
& Accuracy
& Precision
& Recall
& F1-score \\
\midrule
FedAvg
& $87.55{\pm}2.38$
& $87.96{\pm}2.26$
& $87.55{\pm}2.38$
& $87.46{\pm}2.39$ \\

FixedQNN LocalOnly
& $90.31{\pm}1.01$
& $90.54{\pm}0.88$
& $90.31{\pm}1.01$
& $90.25{\pm}1.04$ \\

vFedProtoQNAS w/o QNAS
& $90.01{\pm}0.76$
& $90.21{\pm}0.71$
& $90.01{\pm}0.76$
& $89.92{\pm}0.83$ \\

RandomQNAS + FedProto
& $90.17{\pm}0.92$
& $90.32{\pm}0.87$
& $90.17{\pm}0.92$
& $90.15{\pm}0.90$ \\

PersonalQNAS
& $90.45{\pm}0.70$
& $90.59{\pm}0.66$
& $90.45{\pm}0.70$
& $90.44{\pm}0.71$ \\

\textbf{vFedProtoQNAS (Ours)}
& $\mathbf{91.25{\pm}0.74}$
& $\mathbf{91.29{\pm}0.71}$
& $\mathbf{91.25{\pm}0.74}$
& $\mathbf{91.24{\pm}0.74}$ \\
\bottomrule
\end{tabular}
\vspace{-3mm}
\end{table}

\BfPara{Sensitivity Analysis for Non-IID Strength} Table~\ref{tab:alpha_sensitivity} reports the sensitivity of vFedProtoQNAS to the non-IID strength. The non-IID strength is controlled by the Dirichlet parameter $\alpha$. Larger values of $\alpha$ indicate milder client heterogeneity. For $\alpha=1.0$ and $\alpha=10.0$, vFedProtoQNAS achieves accuracies of $91.15\pm0.76\%$ and $91.16\pm0.93\%$, respectively. These results show that vFedProtoQNAS maintains stable performance under moderate and mild non-IID settings. This suggests that prototype-based collaboration remains effective when class distributions are not extremely imbalanced. In contrast, when $\alpha=0.5$, the accuracy decreases to $84.32\pm11.77\%$. The standard deviation also increases substantially. This result indicates that severe class imbalance makes prototype estimation more sensitive to the data split. In particular, some clients may have insufficient samples for certain classes. This can reduce the reliability of class-wise local prototypes. Overall, vFedProtoQNAS operates stably under typical non-IID settings. However, local prototype reliability becomes a key limiting factor under extreme heterogeneity.

\BfPara{Baseline Comparison} Table~\ref{tab:main_comparison} compares vFedProtoQNAS with the main baselines. The comparison is conducted under the default settings $\beta_{\mathrm{proto}}=0.10$ and $\alpha=1.0$. FedAvg uses the same QNN architecture across all clients. It aggregates local models through parameter averaging. FixedQNN LocalOnly trains a fixed QNN independently on each client. vFedProtoQNAS w/o QNAS and RandomQNAS + FedProto evaluate the effect of prototype sharing. The former removes architecture search and uses a fixed architecture. The latter uses a randomly selected architecture with prototype sharing. PersonalQNAS performs client-specific QNAS without federated prototype sharing. Thus, it evaluates the effect of local personalization alone. vFedProtoQNAS achieves the highest performance among all methods. It obtains an accuracy of $91.25\pm0.74\%$ and an F1-score of $91.24\pm0.74\%$. FedAvg achieves an accuracy of $87.55\pm2.38\%$. This result suggests that parameter averaging is limited in heterogeneous QFL settings. The limitation arises because all clients are forced to use the same architecture. FixedQNN LocalOnly and PersonalQNAS achieve accuracies of $90.31\pm1.01\%$ and $90.45\pm0.70\%$, respectively. These results show that local specialization is important for heterogeneous clients. However, their performance remains lower than that of vFedProtoQNAS. This indicates that local personalization alone does not fully exploit federated collaboration. vFedProtoQNAS w/o QNAS and RandomQNAS + FedProto achieve accuracies of $90.01\pm0.76\%$ and $90.17\pm0.92\%$, respectively. These results show that prototype sharing is effective even without optimized architectures. However, their performance is lower than that of vFedProtoQNAS. This indicates that prototype sharing and personalized QNAS provide complementary benefits. Overall, Table~\ref{tab:main_comparison} shows that vFedProtoQNAS benefits from both components. Personalized QNAS improves client-specific architectural adaptation. Prototype-guided collaboration further improves inter-client representation sharing.

\section{Conclusion}
This paper presented vFedProtoQNAS, a prototype-guided personalized QNAS framework for heterogeneous QFL under client-specific data distributions and device constraints. Unlike conventional QFL approaches that require clients to employ an identical QNN architecture, vFedProtoQNAS allows each client to independently search for and train a personalized architecture. To support collaboration among structurally different QNNs, the proposed framework replaces parameter aggregation with class-wise prototype sharing. The aggregated global prototypes serve as semantic anchors for refining client-specific representations while preserving personalized architectures and local specialization. Experimental results demonstrated that vFedProtoQNAS outperforms conventional parameter-averaging, local-only training, and personalized QNAS baselines. The ablation results further confirmed that personalized architecture search and prototype-guided collaboration provide complementary benefits. Personalized QNAS enables architectural adaptation to local data and device capabilities, while prototype sharing transfers class-level semantic knowledge across heterogeneous client models. In addition, the prototype margin analysis showed that global prototype guidance improves class-consistent representation alignment by encouraging local representations to become closer to the correct class prototype and more distinguishable from incorrect ones. Overall, vFedProtoQNAS provides an effective representation-level collaboration strategy for heterogeneous QFL without requiring structurally aligned client models. This demonstrates prototype-based knowledge sharing can reconcile client-specific architectural personalization with federated collaboration.

\BfPara{Acknowledgement}
The views expressed in this article are those of the authors and do not represent the views of Wells Fargo. This article is for informational purposes only. Nothing contained in this article should be construed as investment advice. Wells Fargo makes no express or implied warranties and expressly disclaims all legal, tax, and accounting implications related to this article. This work was supported by Institute of Information \& Communications Technology Planning \& Evaluation(IITP) grant funded by the Korea government(MSIT) (RS-2024-00439803, SW Star Lab) for Quantum AI Empowered Second-Life Platform Technology

\begingroup
\bibliography{ref}

\bibliographystyle{plainnat}
\endgroup







\end{document}